\documentclass{article}

\usepackage{arxiv}

\usepackage{graphicx}%
\usepackage{multirow}%
\usepackage{amsmath,amssymb,amsfonts}%
\usepackage{amsthm}%
\usepackage{mathrsfs}%
\usepackage[title]{appendix}%
\usepackage{xcolor}%
\usepackage{textcomp}%
\usepackage{manyfoot}%
\usepackage{booktabs}%
\usepackage{algorithmic}%
\usepackage{listings}%
\usepackage[ruled]{algorithm2e}

\usepackage{orcidlink}
\usepackage{tabularx}
\usepackage{hyperref}

\usepackage{lscape}

\hypersetup{
  colorlinks=false,
  linkbordercolor=white,
  urlbordercolor=white,
  pdfborder={0 0 0}
}

\title{Latent Anomaly Detection Through Density Matrices
\thanks{\textit{\underline{Citation}}: 
\textbf{Joseph et al., Latent Anomaly Detection Through Density Matrices}} 
}

\author{
  Joseph A. Gallego-Mejia \orcidlink{0000-0001-8971-4998}, Oscar A. Bustos-Brinez \orcidlink{0000-0003-0704-9117} , Fabio A. Gonz\'{a}lez\orcidlink{0000-0001-9009-7288} \\
  MindLab \\
  Universidad Nacional de Colombia \\
  Bogot\'{a}, Colombia\\
  \texttt{\{jagallegom,oabustosb,fagonzalezo\}@unal.edu.co} 
}

\begin{document}
\maketitle

\begin{abstract}
This paper introduces a novel anomaly detection framework that combines the robust statistical principles of density-estimation-based anomaly detection methods with the representation-learning capabilities of deep learning models. The method originated from this framework is presented in two different versions: a shallow approach employing a density-estimation model based on adaptive Fourier features and density matrices, and a deep approach that integrates an autoencoder to learn a low-dimensional representation of the data. By estimating the density of new samples, both methods are able to find normality scores. The methods can be seamlessly integrated into an end-to-end architecture and optimized using gradient-based optimization techniques. To evaluate their performance, extensive experiments were conducted on various benchmark datasets. The results demonstrate that both versions of the method can achieve comparable or superior performance when compared to other state-of-the-art methods. Notably, the shallow approach performs better on datasets with fewer dimensions, while the autoencoder-based approach shows improved performance on datasets with higher dimensions.
\end{abstract}

\vspace{8pt}


\maketitle

\section{Introduction}

Anomaly detection refers to the task of discovering patterns in the data that deviate from a certain expected behavior \cite{aggarwal2016outlier}. It plays a crucial role in various real-life applications, including but not limited to, identifying fraudulent activities \cite{rizvi2020detection}, monitoring video footage \cite{Wang22video}, detecting defects in industrial processes \cite{denkena2020}, analyzing medical images \cite{medicalanomaly}, and body behaviors \cite{biedebach2023anomaly}. It can also be used as an explainable machine learning algorithm \cite{li2023explainable}. Moreover, it can be employed as a preliminary step in machine learning systems. In these contexts, the main objective is to identify whether a new sample is anomalous or not and utilize this information to make informed decisions \cite{Ruff2021}.  The terminology for anomalies may vary across different domains, encompassing terms like outliers, novelties, exceptions, peculiarities, contaminants, and others. Conventional approaches for anomaly detection have been widely utilized in various applications and knowledge areas \cite{breunig2000lof, scholkopf2001estimating, nachman2020anomaly}, although they lack the representation learning capability found in deep learning models, which enable handling complex, high-dimensional data, such as images \cite{Liu2020}. Multiple deep learning methods, such as variational autoencoders (VAE) \cite{Pol19}, LAKE method \cite{Lv2020}, the deep support vector data description \cite{Ruff2018}, and ALAD, have been introduced in recent years. However, most of them do not produce an explicit density estimation or cannot be trained end-to-end. 

In this study, we propose an anomaly detection method that leverages the probabilistic structure of the data through the use of density matrices, that are based on a quantum mechanics formalism. Our method can be trained end-to-end, thus overcoming the limitations of classical algorithms. The method is presented in two different versions named Anomaly Detection Through Density Matrices (ADDM) and its latent version called LADDM. The latter utilizes the deep representations obtained by an autoencoder. The method starts with either the original representation of the data or the representation given by the autoencoder. These are then passed to an adaptive Fourier feature layer that maps the embedding space to an inner product space in a way that approximates the kernel trick. In this new inner product space, a density matrix is computed as the tensor product summation of the training data points. This method produces density estimates that can be used as normality scores, from which new data points can be labeled as anomalies by using a discriminant threshold.

The contributions of the present work are: 

\begin{itemize}
    \item A \textbf{framework} that combines density matrices with kernel methods to produce normality likelihood estimation.
    \item \textbf{Two new proposals for anomaly detection} which can be trained end-to-end with current deep-learning tools.
    \item A \textbf{systematic comparison} on multiple datasets of the proposals against state-of-the-art anomaly detection methods.
    \item An \textbf{ablation study} was performed to assess the contributions of specific parts of the algorithms.
\end{itemize}

The structure of this paper is as follows. Section \ref{section:background} presents the state of the art in anomaly detection giving a brief summary of representative methods in the field, which are used as a baseline to compare the performance of the proposed methods. Section \ref{section:addm} describes the framework and its main components, presenting ADDM (the shallow approach) and LADDM (the deep approach). Section \ref{section:experiments} defines the experimental evaluation of both versions, discusses the results, and presents an ablation study on LADDM. Finally, Section \ref{section:conclusion} states the conclusions of this work.\\

\textbf{Reproducibility}: all the code of the methods is available at the Github repo https://github.com/Joaggi/lean-dmkde and a release in Zenodo.

\section{Background and Related Work \label{section:background}}

Anomaly detection has been extensively explored in the field of machine learning. The primary objective of the methods that perform it is to identify unusual patterns that deviate from the expected behavior. When the proportion of anomalous data points exceeds 10\%, classification-based approaches like One-Class Support Vector Machines \cite{scholkopf2001estimating} are typically employed. However, in most cases, anomalous data points are scarce, resulting in imbalanced datasets where the proportion of anomalies is usually less than 2\%. In recent years, deep learning algorithms have gained prominence over traditional models, showcasing remarkable performance in anomaly detection tasks \cite{Zhang2021}. These deep algorithms excel in capturing complex data features and conducting feature engineering, particularly when abundant data is available for processing \cite{Rippel2021}. Consequently, they are widely utilized in scenarios involving large datasets with high-dimensional instances, such as images and videos.

\subsection{Baseline Anomaly Detection Methods}

Anomaly detection has been a well-known problem for several centuries. In the half last century, several methods have been proposed to solve general and particular issues in the anomaly sphere. We separated them into three categories: classical methods, recent shallow methods, and deep learning methods. 

Conventional approaches for anomaly detection can be categorized into three primary groups: classification-based methods, such as one-class SVM \cite{scholkopf2001estimating}; distance-based techniques, such as local anomaly factor (LOF) \cite{breunig2000lof} and isolation forest \cite{liu2008isolation}; and statistically-based methods, such as kernel density estimation \cite{nachman2020anomaly}. While these methods have been widely utilized in various applications, they lack the representation learning capability found in deep learning models, which enable handling complex, high-dimensional data, such as images \cite{Liu2020}. Recently, shallow methods like Stochastic Outlier Selection (SOS) \cite{Janssens2012} and COPOD \cite{Li2020copod} have emerged, employing new approaches to these three groups, like affinity and copulas, respectively. However, SOS constructs a quadratic matrix regarding data points, and COPOD does not directly compute joint distributions, which limits its effectiveness. On the other hand, deep learning methods, such as variational autoencoders (VAE) \cite{Pol19}, have also been introduced, although they do not directly address the anomaly detection problem but instead focus on minimizing the reconstruction error as an indirect task. VAE assumes that anomalous points exhibit higher reconstruction errors compared to normal points, but the algorithm has been observed to be capable of adequately reconstructing abnormal points as well. Another deep learning approach, the deep support vector data description \cite{Ruff2018}, encompasses the normal points within a hypersphere but lacks a measure of abnormality. The LAKE method \cite{Lv2020} combines variational autoencoders and kernel density estimation (KDE). However, this method requires saving each latent space in the training dataset for KDE, resulting in significant memory requirements. Lastly, ALAD \cite{zenati2018adversarially} uses the representation captured by a generative adversarial network and compares it against the original data point. 

\section{Anomaly Detection Through Density Matrices: ADDM and LADDM} \label{sec:addm}

The proposed anomaly detection framework gives origin to two algorithms, that can be seen as versions of a single method: Anomaly Detection through Density Matrices (ADDM) and Latent Anomaly Detection through Density Matrices (LADDM). The general model is composed of an autoencoder (in the case of LADDM), an adaptive Fourier feature layer, a quantum measurement layer, and finally a threshold anomaly detection layer. The autoencoder is responsible for performing a dimensionality reduction and calculating the reconstruction error (only for LADDM). The adaptive Fourier feature layer performs a transformation of the data to a Hilbert feature space where the dot product approximates a Gaussian kernel; this mapping can be seen as an optimization from the original formulation of Random Fourier features presented in \cite{rahimi2007rff}. After the mapping, the data points are used to build a density matrix that stores the information from these points and can be seen as an approximation of the probability density function of the data. The quantum measurement layer then uses the density matrix to produce density estimates of the data points. The last step of the algorithm is an anomaly detector that uses the density estimates to find a suitable threshold and classifies each sample as anomalous or normal comparing its density with such threshold. The method depends on two key hyperparameters: the bandwidth of the approximated Gaussian kernel (denoted as $\gamma$) and the size of the adaptive Fourier feature space (denoted as $D$). Each one of the stages of the method is explained in depth in the following sections, with a specific focus on the mathematical foundations of each stage and the transformations performed on the data.

\begin{figure*}[ht]
\begin{centering}
\includegraphics[scale=0.52]{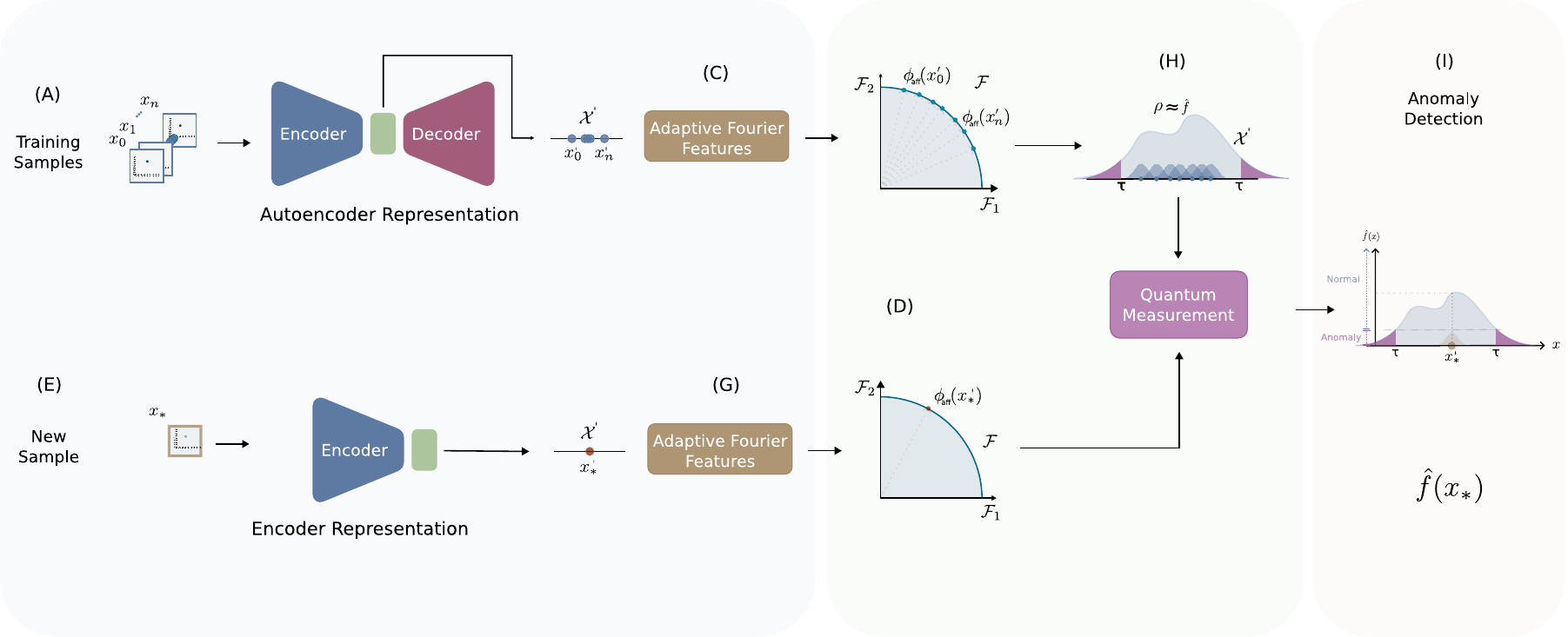}
\par\end{centering}
\caption{Structure of the proposed method, in its deep version (LADDM). Step 1, autoencoder representation learning. Step 2, training of the model using both the reconstruction error and the maximum likelihood estimation of the density matrix $\rho$. Step 3, estimation of the density of a new sample using the learn density matrix $\rho=V^T\Lambda V$. Step 4, anomaly detector using the proportion of the anomalies. The shallow version (ADDM) presents the same structure, except for the first Step.  
\label{fig:model}}
\end{figure*}

The architecture of the proposed framework is shown in Figure \ref{fig:model}.

\begin{table}[h]
    \caption{Notation used throughout this article.}
    \begin{tabularx}
      {\textwidth}{p{0.071\textwidth}XX}      
      \multicolumn{2}{X}{\text{\space \space Mathematical symbols: }} \\
      $\textbf{x}_{1}\cdots\textbf{x}_n$  & set of identical and independent distributed variables\\
      $O(\cdot)$ & complexity in Bachmann Landau notation \\
      $Pr[]$ & probability density function \\
      $k(\cdot, \cdot)$ & kernel function \\
      $\gamma$ & parameter of the Gaussian kernel \\
     $\langle\cdot,\cdot\rangle$ & inner product\\
     $\mathbb{R}^d$ & $d$-dimensional real space \\
     $\mathcal{N}(\cdot)$ & normal distribution \\
     $Uniform$ & uniform distribution \\
     $\rho$ & density matrix \\
     $\Lambda$  & eigenvalues of density matrix \\
     $V$ & eigenvectors of density matrix \\
     $\phi_{\text{aff}}(\cdot)$ & adaptive Fourier feature explicit projection \\
   \end{tabularx}
\end{table}

\subsection{Autoencoder}

The first step of the algorithm is a neural network called an autoencoder. (This only applies for LADDM, however; for ADDM, the method starts in Section 3.2). Given an input space $\mathbb{R}^d$, and a latent space $\mathbb{R}^p$, where typically $p \ll d$, a point $\textbf{x}_i$ in the input space is sent through an $\psi$-encoder and a $\theta$-decoder step, where $\psi: \mathbb{R}^d \rightarrow \mathbb{R}^p$ and $\theta: \mathbb{R}^p \rightarrow \mathbb{R}^d$,  such that:
$$\textbf{z}_i = \psi(\textbf{x}_i, \textbf{w}_\psi)$$
$$\hat{\textbf{x}}_i = \theta(\textbf{z}_i, \textbf{w}_\theta)$$

where $\textbf{w}_\psi$ and $\textbf{w}_\theta$ are the network parameters of the encoder and the decoder respectively, and $\hat{\textbf{x}}_i$ is the reconstruction of the original data. The reconstruction error of the autoencoder is defined as the Euclidean distance between the original data point and its reconstruction:
\begin{align} \label{eq:encoder_loss}
\begin{aligned}
l_{\text{Euc\_dist},i}(\psi, \theta) = -||\textbf{x}_i - \hat{\textbf{x}_i}||^2
\end{aligned}
\end{align}

To capture further information computed by the autoencoder, the Euclidean distance error ($l_{\text{Euc\_dist},i}$) is combined with the cosine similarity ($l_{\text{cos\_sim},i}$), defined as the cosine of the angle between $\textbf{x}_i$ and $\hat{\textbf{x}_i}$, into one vector with the latent space ($\textbf{z}_i$). Such vector, noted as $\textbf{o}_i$, serves as the output of the autoencoder to the next steps of the method, and is given by:

$$\textbf{o}_i = [\textbf{z}_i, l_{\text{Euc\_dist},i}, l_{\text{cos\_sim},i}]$$

\subsection{Fourier Features \label{subsect:training_strategy}}

Random Fourier features were first proposed by \cite{rahimi2007rff}, and the core idea is to build an explicit approximate mapping of the reproducing Hilbert space induced by a given kernel. In particular, if a Gaussian kernel is used, we can approximate $$k(\textbf{x}, \textbf{y}) \simeq \mathbb{E}_{\textbf{w}}(\langle \hat{\phi}_{\text{aff}, \textbf{w}}(\textbf{x}), \hat{\phi}_{\text{aff}, \textbf{w}}(\textbf{y})\rangle )$$
where $\hat{\phi}_{\text{aff}, \textbf{w}} = \sqrt{2} \cos (\textbf{w}^T\textbf{x} + b)$,  $\textbf{w} \sim \mathcal{N}(\textbf{0},\mathbb{I}^d)$ and $\textbf{b} \sim \textit{Uniform}(0, 2\pi)$. This expectation approximation is possible due to Bochner's theorem. 

The adaptive Fourier features are a neural network-based fine-tuning of the random Fourier features. This algorithm, first proposed in \cite{josephQuantumNeural}, selects random pairs of data points $(\textbf{x}_i, \textbf{x}_j)$ and applies gradient descent to find:

$$\textbf{w}^*, b^* = \arg \min_{\textbf{w},b} \frac{1}{m}\sum_{\textbf{x}_i,\textbf{x}_j\in D} (k_\gamma(\textbf{x}_i,\textbf{x}_j) - \hat{k}_{\textbf{w},b}(\textbf{x}_i,\textbf{x}_j))^2$$
where $k_\gamma(\textbf{x}_i,\textbf{x}_j)$ is the Gaussian kernel with a $\gamma$-bandwidth parameter, and $\textbf{w}$ and $b$ are sampled as in Random Fourier features.

\subsection{Quantum Measurement-based Estimation \label{section:addm}}

Density estimation is one of the most studied topics in statistics. Given a random variable $X$ and an arbitrary subset $A$ of its associated measurable space $\mathcal{A}$, the probability density function $f$ is defined as a function whose integral over $A$ equals the probability $Pr[X\in A]$. In general, the underlying density function of an arbitrary system is unknown. 

The most popular method to estimate this density function is kernel density estimation (KDE), also known as Parzen's window. Given a set of training points $\{\textbf{x}_i\}_{i=1,\cdots,N}$, and a point $\textbf{x}$ whose density is to be calculated, KDE calculates it as: 

\begin{equation}
    \hat{f}(\textbf{x})= \frac{1}{N M_\gamma} \sum_{i=1}^N k_\gamma({\textbf{x}_i},\textbf{x})
    \label{eq:kernel_density_estimation}
\end{equation}
where $\gamma$ is the bandwidth parameter, $M\gamma$ is a normalization parameter, and $k_\gamma$ is a kernel with specific properties \cite{girolami03}. KDE is a very precise method, but it has two principal drawbacks: its memory-based approach, i.e., it needs to store each data point $x_1, \cdots, x_n$ in the training phase and uses them to produce inferences in the testing phase, and it cannot be combined with deep learning architectures.  

In \cite{gonzalez2021learning, josephQuantumNeural}, the authors showed that using a derivation of the kernel density estimation method using the Gaussian kernel, we can derive a density estimation method that captures quantum information as follows: 

\begin{align} \label{eq:kde_density_matrix}
\begin{aligned}
\hat{f}(\textbf{x}) & = \frac{1}{M_\gamma} {\phi}_\text{aff}(\textbf{x})^T \rho \: {\phi}_\text{aff}(\textbf{x})
\end{aligned}
\end{align}
where $\rho$ is a density matrix defined as $$ \rho = \frac{1}{N}\sum_{i=1}^N \phi_\text{aff}({\textbf{x}_i})\phi^T_\text{aff}({\textbf{x}_i})$$ and $M_\gamma$ is a normalization constant. Note that $\rho$ is a $D$-dimensional square matrix (its size does not depend on the features of the original data). The density matrix captures the classical and quantum probabilities of a quantum system, and it can be viewed as an aggregation of each data point in the feature space. The $\rho$ matrix is required to be a Hermitian matrix and its trace must be equal to one, $tr(\rho)=1$. \cite{gonzalez2021learning} shows that $\hat{f}$, as defined in Eq. (\ref{eq:kde_density_matrix}), uniformly converges to the Gaussian kernel Parzen's estimator $\hat{f}_{\gamma}$ (\ref{eq:kernel_density_estimation}).

The dimensions of the $\rho$ matrix can be problematic, given its size $D^2$, but this can be alleviated by using the matrix eigen-decomposition $\rho \approx V^T \Lambda V$,
where $V \in \mathcal{R}^{r \times D}$ is a matrix whose rows contain the $r$ vectors with the largest eigenvalues, and $\Lambda \in \mathcal{R}^{r \times r}$ is a diagonal matrix whose values correspond to the $r$ largest eigenvalues of $\rho$. This matrix eigen-decomposition modify the Equation \ref{eq:kde_density_matrix} to:

\begin{align} \label{eq:AFDM_estimation}
\begin{aligned}
\hat{f}_\gamma(\textbf{x})& \simeq \frac{1}{M_\gamma}|| \Lambda^{1/2}V{\phi}_\text{aff}(\textbf{x})||^2
\end{aligned}
\end{align}

Equation \ref{eq:AFDM_estimation} defines the algorithm to perform new density estimates based on density matrices.   The log-likelihood optimization process depends on a loss function $\mathbf{L}$ that can have two different definitions depending on the version of the model. For ADDM, this function is given by:
\begin{equation}
\label{eq:loss_function_addm}
\mathbf{L}_{V^*, \Lambda^*} = \sum_{i=1}^N \log \hat{f}_\gamma (\textbf{x}_i)
\end{equation}

For LADDM, it takes into account the loss function of the autoencoder, as presented in Eq. (\ref{eq:encoder_loss}), and thus becomes:
\begin{equation}
\label{eq:loss_function_lean}
\mathbf{L}_{\textbf{w}_\psi,\textbf{w}_\theta,V^*, \Lambda^*} = (1-\alpha)||\textbf{x}_i - \hat{\textbf{x}_i}||^2 - \alpha \sum_{i=1}^N \log \hat{f}(\textbf{o}_i)
\end{equation}

where the hyperparameter $\alpha$ controls the trade-off between the minimization of the reconstruction error and the maximization of the log-likelihood. In both scenarios, the algorithm is trained to minimize the loss function using gradient descent optimization.

\begin{algorithm}[tb]
\caption{LADDM Training Process}
\label{alg:algorithm}
\textbf{Input}: Training dataset $D=\{\textbf{x}_i\}_{i=1,\cdots,N}$ \\
\textbf{Parameters}: $\alpha$: trade-off between reconstruction error and log-likelihood,\\
$\textbf{w},b$: adaptive Fourier features parameters\\
\textbf{Output}: $\textbf{w}_\psi$: encoder parameters, 
$\textbf{w}_\theta$: decoder parameters, 
$V^*, \Lambda^*$
\begin{algorithmic}[1] 

\STATE Apply gradient descent to find $\textbf{w}^*, b^* = \arg \min_{\textbf{w},b} \frac{1}{m}\sum_{\textbf{x}_i,\textbf{x}_j\in D} (\mathcal{L}(\textbf{x}_i,\textbf{x}_j))^2$ where 
$\mathcal{L}(\textbf{x}_i,\textbf{x}_j) = k_{\gamma}(\psi_{\textbf{w}_\psi}(\textbf{x}_i),\psi_{\textbf{w}_\psi}(\textbf{x}_j))-\hat{k}_{\textbf{w},b}(\psi_{\textbf{w}_\psi}(\textbf{x}_i),\psi_{\textbf{w}_\psi}(\textbf{x}_j))$

\FOR{$\textbf{x}_i \in D \textbf{ until convergence}$}
\STATE $\textbf{z}_i = \psi(\textbf{x}_i, \textbf{w}_\psi)$
\STATE $\hat{\textbf{x}_i} = \theta(\textbf{z}_i, \textbf{w}_\theta)$
\STATE $l_{\text{Euc\_dist},i}=\text{Euclidean\_distance}(\textbf{x}_i,\hat{\textbf{x}}_i)$
\STATE $l_{\text{cos\_sim},i}=\text{cosine\_similarity}(\textbf{x}_i,\hat{\textbf{x}}_i)$
\STATE $\textbf{o}_i = [\textbf{z}_i,l_{\text{Euc\_dist},i}, l_{\text{cos\_sim},i}]$
\STATE $\bar{\phi}_{\mathrm{aff}}(\textbf{z}_i) = \frac{\phi_{\mathrm{aff}}(\textbf{z}_i)}{||\phi_{\mathrm{aff}}(\textbf{z}_i)||}$ where $\phi_{\text{aff}}(\textbf{z}_i)= \sqrt{2}\cos((\textbf{w}^*)^{T} \textbf{z}_i + b^*)$
\STATE  $\hat{f}(\textbf{o}_i) =\frac{1}{M_\gamma}|| \Lambda^{1/2}V\phi_{\text{aff}}(\textbf{z}_i)||^2$  
\STATE $\mathbf{L}_{\textbf{w}_\psi,\textbf{w}_\theta,V^*, \Lambda^*} = (1-\alpha)||\textbf{x}_i - \hat{\textbf{x}_i}||^2 - \alpha \sum_{i=1}^N \log \hat{f}(\textbf{o}_i)$
\STATE $\textbf{w}_\psi,\textbf{w}_\theta,V^*, \Lambda^* \leftarrow \text{update parameters minimizing } \mathbf{L}$

\ENDFOR 

\STATE \textbf{return} $\textbf{w}_\psi,\textbf{w}_\theta,V^*, \Lambda^*$
\end{algorithmic}
\end{algorithm}

\begin{center}
\begin{algorithm}[t]

\caption{ADDM Training Process}
\label{alg:compute_DEMANDE}

\textbf{Input:} Training dataset $T=\{\textbf{x}_i\}_{i=1,\cdots,N}$\\
\textbf{Output:} $V^*, \Lambda^*$

\begin{algorithmic}[1]

\STATE Apply gradient descent to find $\textbf{w}^*, b^* = \arg \min_{\textbf{w},b} \frac{1}{m}\sum_{\textbf{x}_i,\textbf{x}_j\in D} (\mathcal{L}(\textbf{x}_i,\textbf{x}_j))^2$ where 
$\mathcal{L}(\textbf{x}_i,\textbf{x}_j) = k_{\gamma}(\psi_{\textbf{w}_\psi}(\textbf{x}_i),\psi_{\textbf{w}_\psi}(\textbf{x}_j))-\hat{k}_{\textbf{w},b}(\psi_{\textbf{w}_\psi}(\textbf{x}_i),\psi_{\textbf{w}_\psi}(\textbf{x}_j))$
\FOR {$i = 1,\cdots,N$} 
\STATE $\bar{\phi}_{\mathrm{aff}}(\textbf{x}_i) = \frac{\phi_{\mathrm{aff}}(\textbf{x}_i)}{||\phi_{\mathrm{aff}}(\textbf{x}_i)||}$ where $\phi_{\text{aff}}(\textbf{x}_i)= \sqrt{2}\cos((\textbf{w}^*)^{T} \textbf{x}_i + b^*)$
\ENDFOR
\STATE $\rho= \frac{1}{N}\sum_{i=1}^N \bar\phi_\text{aff}(\textbf{x}_i)\bar\phi^T_\text{aff}(\textbf{x}_i)$
\STATE Perform a spectral decomposition of $\rho$ to find $V, \Lambda$ such that $\rho \approx V^T \Lambda V$
\STATE $\text{Apply gradient descent to find }
V^*, \Lambda^* = \arg \min_{V, \Lambda} \sum_{i=1}^N \log \hat{f}(\textbf{x}_i)$\\ where $\hat{f}(\textbf{x}_i) =\frac{1}{M_\gamma}|| \Lambda^{1/2}V\bar\phi_\text{aff}(\textbf{x})||^2$
\STATE \textbf{return} $V^*, \Lambda^*$
\end{algorithmic}
\end{algorithm}
\end{center}

\vspace{5pt}
\subsection{Anomaly Detection Step}

The last step of the anomaly detection algorithm is to detect whether a given data point is a normal or anomalous point. The algorithm uses the proportion of anomalies ($r\%$) within the dataset to obtain a threshold value $\tau$ as:

$$\tau = q_{(r\%)} (\hat{f}(\textbf{x}_1), \cdots, \hat{f}(\textbf{x}_n))$$

where $q$ is the percentile function. In the detection phase, the algorithm uses the $\tau$ value to classify anomalous points as:  
$$\hat{y}(x_{i}) = \left\{\begin{matrix}
 \text{ `normal'} & \textit{if } \hat{f}(x_{i}) \geq \tau\\ 
 \text{ `anomaly'} & \textit{otherwise}
\end{matrix}\right.$$

\begin{algorithm}[t]
\caption{ADDM and LADDM Prediction Process}
\label{alg:prediction_DEMANDE}

\textbf{Input:} Test data point $\{\textbf{x}_*\}, \{\textbf{w}, \textbf{b}\}$ AFF parameters, $\{\Lambda^*, V^* \}$ as density matrix, $r\%$ proportion of anomalies\\
\textbf{Output:} $\hat{f}_\gamma(\textbf{x}_*), \hat{y}(\textbf{x}_*), \tau$
\begin{algorithmic}[1]
\STATE Estimate the density of $\{\textbf{x}_*\}$ as $ \hat{f}_\gamma(\textbf{x}_*)  \simeq \frac{1}{M_\gamma}|| (\Lambda^{*})^{1/2}V^*\bar\phi_\textit{aff}(\textbf{x}_*)||^2$

\STATE Find the threshold as a quantile: $\tau =  q_{(\textit{r\%})} (\hat{F}_{val})$\\ where $\hat{F}_{val}$ is the set of the estimated densities of all test points 
\IF{$\hat{f}_\gamma(\textbf{x}_*) \leq \tau$} 
\STATE $\hat{y}(\textbf{x}_*) = \textit{`anomaly'}$ 
\ELSE
\STATE $\hat{y}(\textbf{x}_*) = \textit{`normal'}$ \ENDIF
\STATE \textbf{return} $\hat{f}_\gamma(\textbf{x}_*), \hat{y}(\textbf{x}_*), \tau$

\end{algorithmic}
\end{algorithm}

\subsection{Complexity Analysis}\label{subsec:threshold_calculation}

\begin{table}[!ht]
    \caption{Complexity analysis of the method, with and without spectral decomposition.}
    \label{table:complexity_analysis}
    \centering
    \small
    \begin{tabular}{|l|l|l|}
    \hline
        ~ & Training & Testing \\ \hline
        Without &  $O(nDd^2+ nD^2)$ & $O(mDd^2 + mD^2)$ \\ \hline
        With & $O(nDd^2 + nD^2 + D^3)$ & $O(mDd^2 + mDr)$ \\ \hline
    \end{tabular}
\end{table}

Table \ref{table:complexity_analysis} shows the complexity analysis with and without spectral decomposition. In this subsection, we use $n$ and $m$ to denote the number of training and testing data points, respectively, while $D$ represents the number of adaptive Fourier features, and $d$ represents the number of features. During the training phase of the algorithm without decomposition, the time complexity is dominated by the dot product calculation between each data point and the adaptive Fourier features. Specifically, the time complexity is proportional to $O(nDd^2)$. Additionally, a dot product is calculated between the resulting vector and the density matrix, with a time complexity proportional to $O(nD^2)$. This gives an overall training time complexity of $O(nDd^2 + nD^2)$. During the testing phase, the time complexity is determined by the dot product calculation between each data point and the adaptive Fourier features, as well as the time complexity of the quantum measurement. Therefore, the overall testing time complexity is proportional to $O(mDd^2 + mD^2)$. For the decomposition variant, the training and testing time complexity is $O(nDd^2 + nD^2 + D^3)$ and $O(mDd^2 + mDr)$, respectively. The memory footprint is determined by the need to maintain a density matrix, whose size is proportional to $O(D^2)$. Additionally, it stores the weights of the adaptive Fourier features, whose size is proportional to $O(dD)$.

\section{Experimental Evaluation \label{section:experiments}}

In this section, we conducted a comprehensive assessment of the proposed anomaly detection algorithms, ADDM and LADDM. To evaluate their performance, we compared them against fourteen diverse anomaly detection algorithms. Our evaluation involved twenty publicly available datasets, varying in size from less than 1,000 values to over 100,000 values. Additionally, these datasets exhibited a wide range of features (dimensions), with some having less than 10 features and others having more than 200. Our dataset selection aimed to encompass a variety of characteristics to ensure the diversity of the datasets used.

To address our research questions, we designed and executed a carefully planned experimental setup. The setup was specifically tailored to provide answers to the following inquiries:

\begin{itemize}
\item \textbf{Q1. Anomaly Detection Comparison:} How accurately do our methods perform in detecting anomalies across a wide range of anomaly detection datasets when compared to state-of-the-art baseline methods?
\item \textbf{Q2. Adaptability:} How well do our methods adapt to datasets with high dimensionality?
\item \textbf{Q3. Ablation Study:} What are the effects on the overall performance of our method when certain stages of the algorithms (specifically, the autoencoder, the reconstruction error as a feature, or the density matrix) are removed?
\end{itemize}

\subsection{Experimental Setup}

The framework used in this paper intends to compare ADDM and LADDM with several baseline algorithms listed in the next subsection. To run One Class SVM, Minimum Covariance Determinant, Local Outlier Factor, and Isolation Forest, the implementation used is the one provided by Scikit-Learn Python library. KNN, SOS, COPOD, LODA, VAE, and DeepSVDD were run using the implementation provided by PyOD Python library \cite{Zhao2019}. LAKE algorithm was implemented based on the Github repository of its authors, although we had to correct the way the test dataset was split to include both normal and anomalous samples in it.

In order to handle the inherent randomness found in some of the algorithms, it was decided to fix in advance (into a single, invariant number) all the random seeds that could affect the different stages of each algorithm. All experiments were carried out on a machine with a 2.1GHz Intel Xeon 64-Core processor with 128GB RAM and two RTX A5000 graphic processing units, that run Ubuntu 20.04.2 operating system.

\subsection{Baseline Methods}
Twelve methods were selected in this setup and are briefly explained next.
\begin{itemize}
    \item Isolation Forest \cite{liu2008isolation}. This technique identifies anomalous data points by constructing lines perpendicular to the origin and measuring the number of tree splits required for each point. The anomalous points are those that require fewer splits. 
    \item Minimum Covariance Determinant \cite{rousseeuw1999fast}. This algorithm fits the data points into a hyperellipsoid using robust estimators of the mean and covariance. Anomalies are detected based on the Mahalanobis distance of each point.
    \item One Class Support Vector Machines \cite{scholkopf2001estimating}. This method creates a maximum-margin hyperplane from the origin, encompassing all training data points. Anomalies are assumed to lie outside this hyperplane, utilizing a Gaussian kernel for the experiments.
    \item Local Outlier Factor \cite{breunig2000lof}. This approach calculates the $k$-nearest neighbors for each point and evaluates the local reachability density to determine the local outlier factor. Points with a LOF value greater than 1 are considered outliers. 
    \item K-nearest neighbors (KNN) \cite{Ramaswamy2000}: This method computes the $k$-nearest neighbor distances for each point and applies a function based on these distances. The selection of an anomaly threshold depends on the chosen distance function.
    \item SOS \cite{Janssens2012}. It is inspired by t-SNE, this algorithm computes a dissimilarity matrix and transforms it into an affinity matrix using a perplexity parameter. By calculating link probabilities between points in sub-graphs, a probability metric for outlier points is obtained. A threshold is then applied for classifying anomalous points.
    \item COPOD \cite{Li2020copod}: This algorithm utilizes copulas, allowing the expression of joint multivariate distribution functions in terms of marginal distribution functions. The copula and a skewness correction are computed to determine if the left or right tail should be used. A threshold is employed for classifying abnormal data points.
    \item LODA \cite{Pevny2016} that is an ensemble algorithm that applies histograms to each dimension of the data. The histograms are projected using random parameters, assuming independence between the projected vectors. A logarithm function is applied to mitigate the curse of dimensionality. Classifications of anomalous points are obtained from the average negative log probability using the histogram projections.
    \item VAE-Bayes \cite{an2015variational}. This model incorporates Bayesian statistics and utilizes a stochastic variational autoencoder for sample reconstruction. Anomalies are determined by high reconstruction errors.
    \item DeepSVDD \cite{Ruff2018}. Based on Support Vector Data Description (SVDD), this method surrounds normal points with a hypersphere in a reproducing Hilbert space. Anomalous points are expected to reside outside the hypersphere, with the parameter $v$ controlling the proportion of anomalies detected. 
    \item  LAKE \cite{Lv2020}: combines two anomaly detection methods. First, a variational autoencoder (VAE) serves as a dimensionality reduction algorithm and measures reconstruction error. Second, a kernel density estimation (KDE) is connected to the last hidden layer of the VAE, incorporating reconstruction error and dimensionality reduction to estimate the density of a given point. 
    \item Adversarially Learned Anomaly Detection (ALAD) \cite{zenati2018adversarially}. is a method that combines a generative adversarial network (GAN) framework with anomaly detection. ALAD leverages adversarial training to learn a representation of normal data and effectively detect anomalies that deviate significantly from the learned distribution. A threshold is then set on these scores to differentiate between normal and anomalous samples. 
    
\end{itemize}

\subsubsection{Datasets}

To evaluate the performance of LADDM for anomaly detection tasks, twenty public datasets were selected. These datasets came from two main sources: the Github repository associated with LAKE\footnote{https://github.com/1246170471/LAKE}, and the ODDS virtual library of Stony Brook University\footnote{http://odds.cs.stonybrook.edu/about-odds/}. The main characteristics of the selected datasets, including their sizes, dimensions, and outlier rates can be seen in Table \ref{dataset_features}. These datasets were chosen because of the wide variety of features they represent, with which the performance of the algorithms can be tested in multiple scenarios, their extensive use in anomaly detection literature, and because of their accessibility, since the files associated with each dataset can be easily accessed in their respective sources. 

\begin{table}[t]

\caption{Main features of the datasets.}
\label{dataset_features}
\centering
\scriptsize
\begin{tabular}[c]{l|c|c|c}
\hline
Dataset & Instances & Dimensions & Outlier Rate \\ 
\hline
Annthyroid & 7200 & 6 & 0,0742 \\
Arrhythmia & 452 & 274 & 0,146 \\
Breastw & 683 & 9 & 0,35 \\
Cardio & 1831 & 21 & 0,096 \\
ForestCover & 286048 & 10 & 0,0096 \\
Glass & 214 & 9 & 0,042 \\
Ionosphere & 351 & 33 & 0,359 \\
Letter & 1600 & 32 & 0,0625 \\
Mammography & 11183 & 6 & 0,02325 \\
MNIST & 7603 & 100 & 0,092 \\
Musk & 3062 & 166 & 0,0317 \\
OptDigits & 5216 & 64 & 0,0288 \\
PenDigits & 6870 & 16 & 0,0227 \\
Pima & 768 & 8 & 0,349 \\
Satellite & 6435 & 36 & 0,3164 \\
SatImage & 5803 & 36 & 0,0122 \\
Shuttle & 49097 & 9 & 0,0715 \\
SpamBase & 3485 & 58 & 0,2 \\
Speech & 3686 & 400 & 0,01655 \\
Thyroid & 3772 & 6 & 0,025 \\
Vertebral & 240 & 6 & 0,125 \\
Vowels & 1456 & 12 & 0,03434 \\
WBC & 378 & 30 & 0,0556 \\
Wine & 129 & 13 & 0,07752 \\ 
\hline
\end{tabular}

\end{table}

\subsubsection{Metrics}

To compare the performance of the proposed algorithm against the baseline anomaly detection methods, two different metrics were selected: the area under the Recall-Precision curve (AUC-PR) and the area under the ROC curve (AUC-ROC). These are two metrics commonly used to compare classification algorithms and are especially useful when there is a high imbalance between the different classes, as is the case here. The calculations of the metrics follow the implementation provided in Scikit-Learn.

For each algorithm, a set of parameters of interest was selected to perform a series of searches for the combinations of parameters that gave the best results for each dataset under study. The list of the parameters under study for each method can be found in Table \ref{parameters}.

\begin{table}[ht!]

\caption{Parameters of all the algorithms selected for a grid search.}
\label{parameters}
\scriptsize
\centering
\begin{tabular}[c]{p{0.21\textwidth}|p{0.5\textwidth}}
\hline
Algorithm & Parameters of Interest \\
\hline

One Class SVM & gamma (Gaussian kernel), outlier percentage \\
Isolation Forest & number of estimators, samples per estimator, outlier percentage \\
Covariance & outlier percentage \\
LOF (Local Outlier Factor) & number of neighbors, outlier percentage \\
K-nearest Neighbors & number of neighbors, outlier percentage \\
SOS & perplexity (matrix parameter), outlier percentage \\
COPOD & outlier percentage \\
LODA & outlier percentage \\
VAE-Bayes & outlier percentage \\
DeepSVDD & outlier percentage \\
LAKE & normality ratio (related to outlier percentage) \\
ADDM & sigma (Gaussian kernel), size of Fourier features mapping, size of density matrix eigen-decomposition\\
LADDM & sigma (Gaussian kernel), architecture of the autoencoder, size of Fourier features mapping, alpha (trade-off parameter)  \\

\hline
\end{tabular}

\end{table}

The selection of the best parameters was made by using a grid search strategy, ranking all the possible combinations of parameters (up to a limit of 100 experiments) in terms of the metrics, and choosing the combination that showed the highest value for AUC-ROC and, in case of a tie, for AUC-PR. This selection of parameters was performed for each algorithm and each dataset.

\subsection{Results and Discussion}

The results for both metrics, obtained when using the winning combinations for all experiments (algorithm and dataset pairs), are reported in Table \ref{table:aucpr_results} for AUC-PR metric and in Table \ref{table:aucroc_results} for AUC-ROC metric. When considering the average performance, there is a noticeable difference between the proposed methods ADDM and LADDM, and the performance of most of the baseline methods, especially for the AUC-ROC metric. In both metrics, our methods show the first and second-best performance, followed by methods like the Autoencoder, Isolation Forest, and KNN, which had a better performance than deep learning alternatives like VAE-Bayes or LAKE.

When considering the dataset features over the performance of the methods, some patterns emerge. Dimensionality does not seem to influence our methods since they perform similarly for datasets with low dimensions (like MNIST or ForestCover) and for datasets with high dimensions (like Glass or Pima). Concerning the outlier rate of the datasets, the methods perform well for datasets whose outlier rate lies below 10\% or above 30\% (like Optdigits or Satellite), but not so much with datasets between 10\% and 30\% (with relatively bad results in datasets like Shuttle or Arrhythmia). There is also a notable relationship with the size of the dataset: LADDM performs slightly better for datasets of sizes below 1000 samples like Vertebral or Wine, and ADDM for datasets with samples between 1000 and 5000 samples, like Musk or Letter. 

\begin{landscape}
\setlength{\tabcolsep}{3pt}
\begin{table}[t!]
\caption{Area under the Precision-Recall curve (AUC-PR) for all algorithms applied on all datasets. The three highest values have been highlighted, the first in bold, the second in underlined and the third in italics.}
\label{table:aucpr_results}
\centering
\begin{small}
\begin{tabular}{l|cc|ccccccccccccr}
    \hline
    \hline
        Dataset & LADDM & ADDM & Autoenc. & iForest & Covar. & KNN & VAE-B & COPOD & LODA & ALAD & OCSVM & LOF & SOS & LAKE & DSVDD \\ \hline
        Annthyroid & \textit{0,248} & 0,187 & 0,232 & \underline{0,34} & \textbf{0,504} & 0,212 & 0,209 & 0,197 & 0,147 & 0,192 & 0,127 & 0,186 & 0,142 & 0,174 & 0,083 \\ 
        Arrhythmia & 0,454 & \underline{0,534} & 0,515 & \textit{0,558} & 0,459 & 0,514 & 0,532 & \textbf{0,566} & 0,365 & 0,393 & 0,425 & 0,433 & 0,472 & 0,116 & 0,32 \\ 
        Breastw & \underline{0,988} & \textit{0,987} & 0,98 & \textbf{0,994} & 0,969 & \underline{0,988} & 0,968 & \textit{0,987} & 0,963 & 0,958 & 0,825 & 0,335 & 0,886 & 0,864 & 0,904 \\ 
        Cardio & 0,47 & \underline{0,627} & \textit{0,604} & \textbf{0,693} & 0,469 & 0,555 & 0,595 & 0,581 & 0,198 & 0,507 & 0,465 & 0,185 & 0,254 & 0,116 & 0,267 \\ 
        ForestCover & \underline{0,465} & \textbf{0,494} & 0,033 & 0,097 & 0,015 & 0,06 & 0,07 & 0,06 & 0,065 & 0,043 & \textit{0,116} & 0,015 & - & 0,095 & 0,01 \\ 
        Glass & \textbf{0,643} & \underline{0,167} & 0,077 & 0,072 & 0,097 & 0,121 & 0,077 & 0,076 & 0,058 & 0,056 & 0,035 & 0,035 & 0,072 & 0,036 & \textit{0,143} \\ 
        Ionosphere & 0,797 & \textbf{0,984} & \underline{0,966} & 0,711 & 0,92 & 0,951 & 0,778 & 0,77 & 0,449 & 0,848 & 0,754 & 0,837 & 0,843 & 0,228 & \textit{0,958} \\ 
        Letter & 0,156 & \textbf{0,489} & \textit{0,29} & 0,111 & 0,241 & 0,228 & 0,087 & 0,072 & 0,063 & 0,081 & 0,075 & \underline{0,375} & 0,288 & 0,036 & 0,271 \\ 
        Mammogr. & 0,253 & 0,187 & \underline{0,356} & \textit{0,307} & 0,145 & 0,193 & 0,196 & \textbf{0,416} & 0,154 & 0,19 & 0,042 & 0,113 & 0,072 & 0,072 & 0,078 \\ 
        MNIST & \textbf{0,71} & \textit{0,549} & \underline{0,551} & 0,326 & 0,494 & 0,468 & 0,431 & 0,246 & 0,185 & 0,264 & 0,216 & 0,309 & 0,228 & 0,333 & 0,148 \\ 
        Musk & 0,747 & \textbf{1} & \textbf{1} & 0,891 & \textit{0,982} & \underline{0,992} & \textbf{1} & 0,449 & 0,545 & 0,142 & 0,214 & 0,032 & 0,251 & 0,853 & 0,026 \\ 
        OptDigits & \textit{0,407} & \underline{0,526} & \textbf{0,555} & 0,034 & 0,021 & 0,022 & 0,028 & 0,044 & 0,017 & 0,021 & 0,024 & 0,032 & 0,032 & 0,021 & 0,026 \\ 
        PenDigits & \textbf{0,672} & \underline{0,614} & 0,173 & \textit{0,339} & 0,092 & 0,208 & 0,208 & 0,162 & 0,289 & 0,121 & 0,118 & 0,018 & 0,063 & 0,06 & 0,048 \\ 
        Pima & \underline{0,572} & \textbf{0,596} & 0,487 & 0,463 & 0,49 & \textit{0,497} & 0,45 & 0,479 & 0,339 & 0,424 & 0,393 & 0,463 & 0,415 & 0,319 & 0,453 \\ 
        Satellite & \underline{0,809} & \textbf{0,882} & \textit{0,795} & 0,66 & 0,76 & 0,61 & 0,631 & 0,601 & 0,596 & 0,561 & 0,54 & 0,36 & 0,354 & 0,23 & 0,32 \\ 
        SatImage & 0,918 & 0,764 & \underline{0,948} & 0,912 & 0,635 & \textbf{0,955} & 0,822 & 0,751 & \textit{0,923} & 0,575 & 0,394 & 0,06 & 0,056 & 0,175 & 0,014 \\ 
        Shuttle & \textbf{0,986} & 0,957 & \textit{0,96} & \underline{0,968} & 0,842 & 0,19 & 0,922 & 0,954 & 0,911 & 0,878 & 0,568 & 0,1 & 0,111 & 0,163 & 0,072 \\ 
        SpamBase & \textbf{0,966} & \underline{0,818} & 0,487 & 0,372 & 0,307 & 0,345 & 0,348 & 0,44 & 0,302 & 0,264 & 0,273 & 0,155 & 0,26 & \textit{0,576} & 0,231 \\ 
        Speech & \underline{0,043} & 0,016 & 0,019 & 0,023 & 0,022 & 0,019 & 0,02 & 0,02 & 0,029 & 0,02 & 0,019 & \textbf{0,044} & 0,024 & 0,026 & \textit{0,028} \\ 
        Thyroid & 0,364 & 0,374 & \textit{0,499} & \underline{0,675} & \textbf{0,688} & 0,324 & 0,442 & 0,223 & 0,131 & 0,402 & 0,286 & 0,293 & 0,082 & 0,013 & 0,125 \\ 
        Vertebral & \textbf{0,787} & \underline{0,26} & 0,197 & 0,133 & 0,138 & 0,125 & 0,165 & 0,119 & 0,106 & 0,123 & \textit{0,218} & 0,153 & 0,121 & 0,164 & 0,077 \\ 
        Vowels & 0,276 & \textbf{0,757} & \textit{0,574} & 0,074 & 0,05 & \underline{0,641} & 0,148 & 0,049 & 0,05 & 0,075 & 0,083 & 0,374 & 0,199 & 0,267 & 0,217 \\ 
        WBC & \underline{0,758} & \textit{0,703} & 0,489 & 0,587 & 0,597 & 0,484 & 0,506 & 0,7 & 0,626 & 0,161 & 0,229 & \textbf{0,76} & 0,476 & 0,237 & 0,192 \\ 
        Wine & \textbf{1} & 0,467 & \underline{0,756} & 0,61 & \underline{0,756} & 0,444 & 0,297 & 0,533 & 0,494 & 0,421 & 0,242 & \textit{0,639} & 0,091 & 0,274 & 0,137 \\ 

    \hline
    \hline
        Mean & \textbf{0,6036}& \underline{0,5807}& \textit{0,5226} & 0,4562 & 0,4455 & 0,4227 & 0,4137 & 0,3956 & 0,3335 & 0,3216 & 0,2783 & 0,2627 & 0,2413 & 0,227 & 0,2145 \\ 

    \hline
    \end{tabular}
\end{small}
\end{table}
\end{landscape}

\begin{landscape}
\begin{table}[t]
\setlength{\tabcolsep}{3pt}
\caption{Area under the ROC curve (AUC-ROC) for all algorithms applied on all datasets. The three highest values have been highlighted, the first in bold, the second in underlined and the third in italics.}
\label{table:aucroc_results}

\centering
\begin{small}
\begin{tabular}{l|cc|ccccccccccccr}
    \hline
    \hline
        Dataset & LADDM & ADDM & Autoenc. & iForest & Covar. & KNN & VAE-B & COPOD & LODA & ALAD & OCSVM & LOF & SOS & LAKE & DSVDD \\ \hline
        Annthyroid & 0,733 & 0,788 & 0,648 & \underline{0,833} & \textbf{0,918} & 0,704 & 0,704 & \textit{0,799} & 0,66 & 0,687 & 0,617 & 0,71 & 0,642 & 0,7 & 0,51 \\ 
        Arrhythmia & 0,753 & 0,773 & 0,778 & \textbf{0,863} & \textit{0,817} & 0,776 & 0,769 & 0,787 & 0,676 & 0,724 & 0,807 & \underline{0,832} & 0,738 & 0,31 & 0,707 \\ 
        Breastw & \underline{0,994} & \underline{0,994} & 0,989 & \textbf{0,997} & 0,985 & \textit{0,993} & 0,961 & \textit{0,993} & 0,962 & 0,969 & 0,899 & 0,518 & 0,927 & 0,93 & 0,946 \\ 
        Cardio & 0,761 & 0,813 & \underline{0,951} & \underline{0,951} & 0,863 & \textit{0,937} & \textbf{0,953} & 0,919 & 0,677 & 0,924 & 0,893 & 0,645 & 0,829 & 0,474 & 0,655 \\ 
        ForestCover & \underline{0,972} & \textbf{0,987} & 0,837 & \textit{0,939} & 0,686 & 0,891 & 0,933 & 0,877 & 0,905 & 0,859 & 0,877 & 0,558 & - & 0,878 & 0,5 \\ 
        Glass & \textit{0,769} & \textbf{0,885} & 0,663 & 0,635 & 0,74 & \underline{0,788} & 0,663 & 0,654 & 0,49 & 0,529 & 0,212 & 0,221 & 0,558 & 0,25 & 0,76 \\ 
        Ionosphere & 0,806 & \textbf{0,988} & \underline{0,978} & 0,772 & 0,916 & 0,964 & 0,833 & 0,853 & 0,456 & 0,874 & 0,743 & 0,842 & 0,84 & 0,069 & \textit{0,965} \\ 
        Letter & 0,637 & \textbf{0,879} & \underline{0,837} & 0,647 & \textit{0,83} & 0,822 & 0,499 & 0,532 & 0,464 & 0,509 & 0,462 & \underline{0,837} & 0,816 & 0,138 & 0,554 \\ 
        Mammogr. & \underline{0,907} & \textbf{0,914} & 0,885 & 0,88 & 0,747 & 0,845 & 0,883 & \textit{0,908} & 0,773 & 0,84 & 0,557 & 0,786 & 0,626 & 0,703 & 0,531 \\ 
        MNIST & \textbf{0,948} & \underline{0,925} & 0,893 & 0,8 & \textit{0,908} & 0,876 & 0,856 & 0,784 & 0,658 & 0,734 & 0,693 & 0,767 & 0,739 & 0,842 & 0,543 \\ 
        Musk & 0,984 & \textbf{1} & \textbf{1} & \textit{0,993} & \underline{0,999} & \textbf{1} & \textbf{1} & 0,958 & 0,938 & 0,717 & 0,813 & 0,442 & 0,856 & 0,881 & 0,408 \\ 
        OptDigits & \textit{0,945} & \textbf{0,991} & \underline{0,975} & 0,568 & 0,355 & 0,394 & 0,522 & 0,694 & 0,215 & 0,368 & 0,424 & 0,417 & 0,548 & 0,33 & 0,433 \\ 
        PenDigits & \underline{0,977} & \textbf{0,996} & 0,91 & 0,941 & 0,862 & \textit{0,96} & 0,939 & 0,907 & 0,925 & 0,896 & 0,865 & 0,374 & 0,683 & 0,65 & 0,516 \\ 
        Pima & \textbf{0,744} & \underline{0,705} & 0,655 & 0,666 & \textit{0,685} & 0,677 & 0,608 & 0,613 & 0,378 & 0,604 & 0,533 & 0,638 & 0,589 & 0,465 & 0,574 \\ 
        Satellite & \underline{0,839} & \textbf{0,875} & \textit{0,827} & 0,674 & 0,79 & 0,758 & 0,625 & 0,658 & 0,598 & 0,609 & 0,582 & 0,542 & 0,563 & 0,294 & 0,498 \\ 
        SatImage & 0,964 & 0,99 & \underline{0,997} & \underline{0,997} & 0,994 & \textbf{0,998} & 0,99 & 0,979 & \textit{0,995} & 0,939 & 0,924 & 0,69 & 0,816 & 0,837 & 0,54 \\ 
        Shuttle & 0,988 & 0,987 & \textit{0,991} & \textbf{0,994} & 0,989 & 0,716 & 0,988 & \underline{0,993} & 0,973 & 0,988 & 0,927 & 0,541 & 0,505 & 0,634 & 0,501 \\ 
        SpamBase & \textbf{0,969} & \underline{0,884} & \textit{0,861} & 0,738 & 0,669 & 0,722 & 0,692 & 0,766 & 0,637 & 0,63 & 0,667 & 0,384 & 0,581 & 0,851 & 0,536 \\ 
        Speech & \textbf{0,709} & 0,513 & 0,504 & 0,483 & 0,516 & 0,507 & 0,505 & 0,534 & 0,562 & 0,582 & 0,495 & \textit{0,597} & 0,568 & 0,531 & \underline{0,651} \\ 
        Thyroid & 0,855 & 0,951 & 0,97 & \underline{0,99} & \textbf{0,993} & 0,945 & 0,958 & 0,937 & 0,585 & \textit{0,972} & 0,929 & 0,951 & 0,741 & 0,075 & 0,605 \\ 
        Vertebral & \textbf{0,965} & 0,528 & \textit{0,593} & 0,483 & 0,498 & 0,477 & 0,566 & 0,477 & 0,383 & 0,501 & \underline{0,69} & 0,524 & 0,45 & 0,523 & 0,143 \\ 
        Vowels & 0,732 & \textbf{0,982} & \textit{0,916} & 0,668 & 0,653 & \underline{0,961} & 0,613 & 0,539 & 0,54 & 0,733 & 0,58 & 0,912 & 0,825 & 0,298 & 0,555 \\ 
        WBC & \underline{0,96} & 0,922 & 0,913 & \textit{0,94} & \textit{0,94} & 0,909 & 0,898 & \textit{0,94} & 0,909 & 0,762 & 0,464 & \textbf{0,976} & 0,918 & 0,371 & 0,633 \\ 
        Wine & \textbf{1} & \underline{0,967} & \underline{0,967} & 0,922 & \underline{0,967} & 0,9 & 0,833 & \textit{0,944} & 0,767 & 0,9 & 0,8 & \underline{0,967} & 0,5 & 0,7 & 0,556 \\ 
    \hline
    \hline
        Mean & \underline{0,8713} & \textbf{0,8848} & \textit{0,8557} & 0,8072 & 0,805 & 0,8133 & 0,7829 & 0,7935 & 0,6719 & 0,7437 & 0,6855 & 0,6529 & 0,6607 & 0,5305 & 0,5758\\ 
 \hline

    \end{tabular}
\end{small}
\end{table}
\end{landscape}

\begin{figure}[h]
    \centering
    \includegraphics[width=1.05\textwidth]{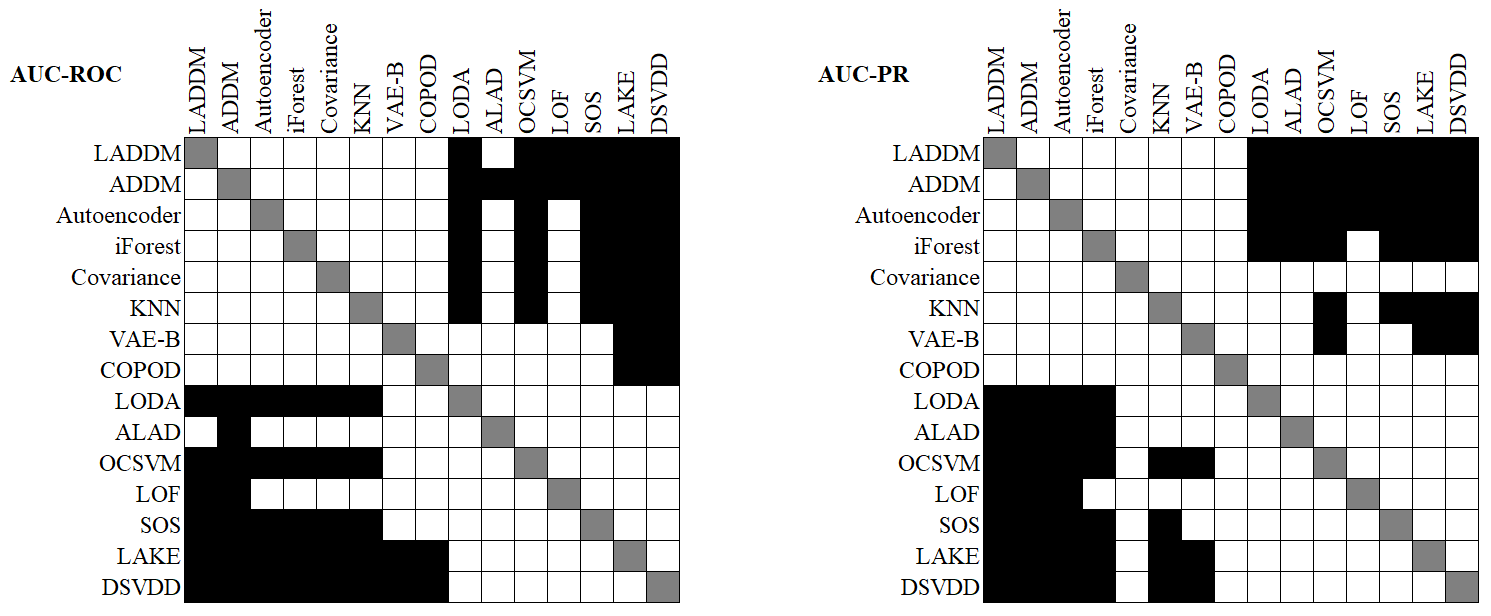}
    \caption{Results for Friedman-Nemenyi test over the metrics. Our methods (LADDM and AD-DMDKE) are the most different from others.}
    \label{fig:chess_nemenyi}
\end{figure}

Over these results, a statistical analysis was performed. In particular, the Friedman test was applied over the rankings of both metrics to conclude that there is a statistically significant difference between the methods; then, a Friedman-Nemenyi test was also applied. The results of this latter test can be seen in Figure \ref{fig:chess_nemenyi}, where black squares correspond to pairs of datasets that differ significantly, and white squares correspond to pairs that do not. AUC-PR (the right grid) shows that the most different methods are LADDM, ADDM, and the Autoencoder, due to their good performances; while AUC-ROC (the left grid) show that the most different methods are LAKE and DeepSVDD, which have a consistently bad performance over the datasets.

\subsection{Ablation Study (for LADDM)}

To establish that LADDM can achieve a better performance than each of its separate components, the following experiments were performed: 

\begin{itemize}
    \item KDE only: density estimation is at the core of the proposed method, so a classic approach to KDE (specifically, the one from Scikit-Learn library) was also applied to the datasets. It was allowed to operate with different kernel functions and bandwidths.
    \item AutoEncoder only: since the proposed method uses an autoencoder as a mechanism to create a latent representation of the data, the neural network alone was applied over the datasets, in such a way that the reconstruction error was used as a measure of anomaly.
    \item Removing reconstruction measure: LADDM uses both reconstruction measure (from the autoencoder) and estimation error (from the quantum estimation stage) as the loss function that it tries to minimize. A LADDM version (called LADDM NoRecon) where only the estimation measure is considered as the loss function was also applied to the datasets.
\end{itemize}

The performance of these experiments over some of the datasets was compared with LADDM using F1-Score as before. A parameter grid search was also performed over the parameters that were applied in each case and with the same value ranges used in LADDM. Results can be seen in Table \ref{ablation}. From these, it is possible to see that LADDM performs effectively better than isolating KDE or the Autoencoder stages, mostly with the latter. KDE shows good performances in some datasets, possibly since it was allowed to work with different kernel shapes, some of which can fit better to certain datasets than the Gaussian kernel that LADDM utilizes. Autoencoder alone and LADDM NoRecon do not seem to perform as well as LADDM (except in a few cases), showing that relying only on reconstruction error is a notable shortcoming when compared to combining it with density estimation.

\setlength{\tabcolsep}{6pt}
\begin{table}[ht!]

\caption{Results (F1-Score) for ablation study of LADDM.}
\label{ablation}

\centering
\scriptsize
\begin{tabular}[c]{l||c|c|c||c}
\hline
Dataset     & KDE   & AE & \tiny{LADDM NoRecon} & LADDM \\
\hline
Arrhythmia  & 0,895 & 0,865 & 0,887 & \bf{0,902} \\
Cardio      & 0,847 & 0,809 & 0,847 & \bf{0,856} \\
Glass       & 0,934 & 0,939 & 0,974 & \bf{1,000} \\
Ionosphere  & 0,946 & 0,814 & 0,959 & \bf{0,973} \\
Letter      & \bf{0,929} & 0,907 & 0,914 & 0,919 \\
MNIST       & 0,925 & 0,883 & 0,918 & \bf{0,930} \\
Musk        & \bf{1,000} & 0,981 & \bf{1,000} & \bf{1,000} \\
OptDigits   & \bf{0,986} & 0,959 & 0,960 & 0,958 \\
PenDigits   & \bf{0,996} & 0,968 & 0,989 & 0,989 \\
Pima        & 0,734 & 0,708 & 0,733 & \bf{0,742} \\
Satellite   & 0,821 & 0,689 & 0,848 & \bf{0,852} \\
SatImage    & 0,998 & 0,992 & \bf{1,000} & \bf{1,000} \\
SpamBase    & 0,872 & 0,787 & 0,942 & \bf{0,947} \\
Thyroid     & 0,962 & 0,954 & \bf{0,973} & 0,972 \\
Vertebral   & 0,827 & 0,776 & 0,855 & \bf{0,870} \\
Vowels      & 0,977 & 0,959 & \bf{0,993} & 0,986 \\
WBC         & 0,950 & 0,931 & 0,961 & \bf{0,972} \\
\hline
\end{tabular}

\end{table}

\vspace{5pt}
\section{Conclusions \label{section:conclusion}}

This paper presented ADDM and LADDM, two versions of a framework for anomaly detection based on the combination of adaptive Fourier features and density estimation through quantum measurements. LADDM also includes an autoencoder to make use of the deep representation capabilities of this neural network. These two versions were compared against eleven different anomaly detection algorithms, using a framework that included twenty labeled anomaly detection datasets. For each dataset and algorithm, a grid search of the best parameters was performed, and the performance of the winning algorithms was compared using F1-Score (weighted average) as the main metric. Both ADDM and LADDM showed state-of-the-art performance, being superior to most classic algorithms and comparable to deep learning-based methods. Their performance does not seem to be affected by the features of the dataset, although ADDM highlights in low-dimensional datasets and LADDM in some high-dimensional datasets. Also, LADDM performs better than its separate parts (KDE and autoencoder) and there is a noticeable difference when using only reconstruction measures, all of which show that the combination of density estimation and reconstruction from autoencoders can perform better than these two elements separately.








\bibliography{references}
\bibliographystyle{splncs04}

\end{document}